%% file: main.tex
\documentclass[conference]{IEEEtran}
\IEEEoverridecommandlockouts
\usepackage{cite}
\usepackage{amsmath,amssymb,amsfonts}
\usepackage{algorithm}
\usepackage{algorithmic}
\usepackage{graphicx}
\usepackage{textcomp}
\usepackage{xcolor}
\usepackage{subfigure}
\usepackage{bbm}
\usepackage{multirow}
\usepackage{amsthm}
\usepackage{pifont}

\allowdisplaybreaks[4]

\newtheorem*{theorem}{Theorem}
\def\BibTeX{{\rm B\kern-.05em{\sc i\kern-.025em b}\kern-.08em
    T\kern-.1667em\lower.7ex\hbox{E}\kern-.125emX}}

\begin{document}

\title{Quality meets Diversity: A Model-Agnostic Framework for Computerized Adaptive Testing}

\author{Haoyang Bi$^1$, Haiping Ma$^2$, Zhenya Huang$^{1,*}$\thanks{* denotes the corresponding author}, Yu Yin$^1$, Qi Liu$^1$, Enhong Chen$^1$, Yu Su$^3$, Shijin Wang$^3$\\
$^1$Anhui Province Key Laboratory of Big Data Analysis and Application \\
School of Computer Science and Technology, University of Science and Technology of China \\ 
\{bhy0521, yxonic\}@mail.ustc.edu.cn,
\{huangzhy, qiliuql, cheneh\}@ustc.edu.cn \\
$^2$Anhui University, hpma@ahu.edu.cn \\
$^3$IFLYTEK Research,  
\{yusu, sjwang\}@iflytek.com}

\maketitle

\begin{abstract}
Computerized Adaptive Testing (CAT) is emerging as a promising testing application in many scenarios, such as education, game and recruitment, which targets at diagnosing the knowledge mastery levels of examinees on required concepts. It shows the advantage of tailoring a personalized testing procedure for each examinee, which selects questions step by step, depending on her performance. While there are many efforts on developing CAT systems, existing solutions generally follow an inflexible \textit{model-specific} fashion. That is, they need to observe a specific cognitive model which can estimate examinee's knowledge levels and design the selection strategy according to the model estimation. In this paper, we study a novel \textit{model-agnostic} CAT problem, where we aim to propose a flexible framework that can adapt to different cognitive models. Meanwhile, this work also figures out CAT solution with addressing the problem of how to generate both high-quality and diverse questions simultaneously, which can give a comprehensive knowledge diagnosis for each examinee. Inspired by Active Learning, we propose a novel framework, namely \textbf{M}odel-\textbf{A}gnostic \textbf{A}daptive \textbf{T}esting (MAAT) for CAT solution, where we design three sophisticated modules including Quality Module, Diversity Module and Importance Module. Specifically, at one CAT selection step, Quality Module first quantifies the informativeness of questions and generates candidate subset with the highest quality. Then, Diversity Module selects one question at each step that maximizes the concept coverage. Additionally, we propose Importance Module to estimate the importance of concepts that optimizes the CAT selection. Under MAAT, we prove that the goal of maximizing both quality and diversity is NP-hard, but we provide efficient algorithms by exploiting the inherent submodular property. Extensive experimental results on two real-world datasets clearly demonstrate that our MAAT can support CAT with guaranteeing both quality and diversity perspectives.

\end{abstract}


\input{introduction}

\input{related-work}

\input{preliminaries}

\input{MAAT}

\input{experiment}

\input{conclusion}

\input{acknowledgement}

\bibliographystyle{IEEEtran}
\bibliography{reference}
\end{document}

%% file: introduction.tex
\section{Introduction}
\label{section:intro}

Designing appropriate tests to evaluate the knowledge states on required concepts of examinees is a fundamental task in many real-world scenarios, such as education, game and job recruit~\cite{minn2018deep,hang2018exploring}. Traditionally, instructors can organize a pencil-paper test, which carefully selects a set of questions for examinees at one time, and therefore we can assess the states of them from their performances. Although such simple way is effective, it just provides all examinees with the same environment so it is difficult to guarantee the rationality of all selected questions~\cite{magis2017computerized}. Therefore, recent efforts focus on another testing form called \textit{Computerized Adaptive Testing} (CAT), which aims to build tests that personally adapt to each examinee, tailoring questions step by step, depending on her performances~\cite{wainer2000computerized}. In fact, CAT has many advantages including improving accuracy, guaranteeing security and enhancing examinee engagement, which has already been applied in many standard test organizations, such as Graduate Management Admission Test (GMAT) \cite{rudner2009implementing} and Graduate Record Examinations (GRE) \cite{mills2000gre}.

\begin{figure}
    \centering
    \subfigure[]{    
        \includegraphics[width=0.45\linewidth]{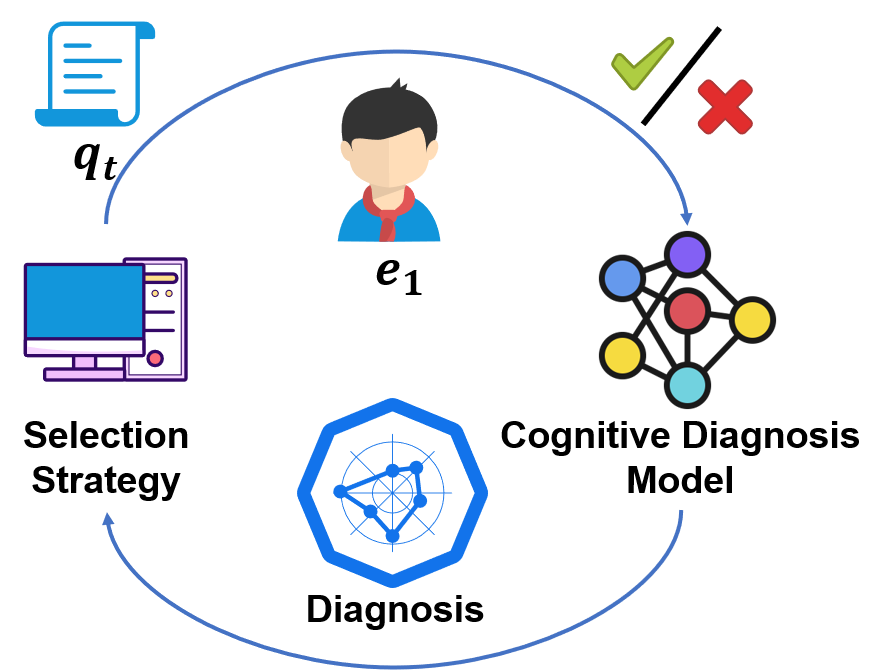}
        \label{subfig:CATcomponents}
    }
    \subfigure[]{    
        \includegraphics[width=0.45\linewidth]{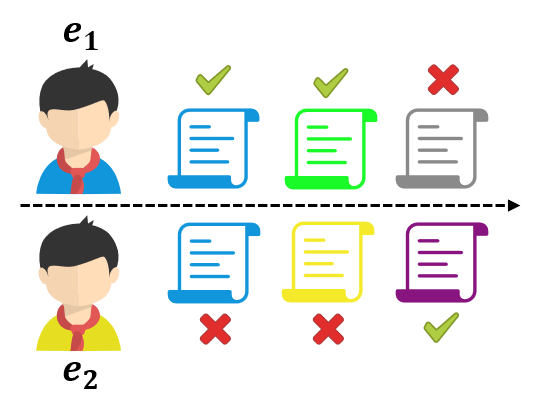}
        \label{subfig:CATprocedure}
    }
    \caption{(a) Illustration of a typical CAT system and its testing procedure in one step. (b) Two toy examples of CAT procedure taken by examinees $e_1$ and $e_2$. We represent different questions with different colors. }
    \label{Fig.1}
    \vspace{-15pt}
\end{figure}

In practice, a typical CAT system generally consists of two key components \cite{van2010elements,magis2017computerized}: (1) a \textit{cognitive diagnosis model} (CDM) that estimates the knowledge states of examinees according to their performance; (2) a \textit{selection strategy} that chooses a question from the pool to support the testing procedure. As shown in Fig.~\ref{subfig:CATcomponents}, when an examinee $e_1$ comes, such CAT system can establish an interactive testing procedure for her. At step $t$, the system first posts one question (e.g., $q_t$). Then, she reads and answers it. After receiving the response (i.e., right or wrong), the system with CDM estimates her current states and on the basis carefully selects a new question $q_{t+1}$ at the next round. This procedure repeats several times until meeting the termination like reaching the maximum testing length~\cite{vie2017review}, so that we can realize how much she has learned about the required concepts (e.g., ``Function'' in Math). In this way, even if starting with the same question, examinees, e.g., $e_1$ and $e_2$ in Fig.~\ref{subfig:CATprocedure}, still can be tailored personalizations. Therefore, the key issue is how to establish an optimal CAT system for choosing the appropriate questions for examinees.

In the literature, there are many efforts on designing CAT, which have already supported many standard tests~\cite{chang2015psychometrics}. Generally, existing solutions deeply dig into underlying CDMs, such as item response theory (IRT)~\cite{embretson2013item} and multidimensional one (MIRT)~\cite{ackerman2003using}, and then produces questions via observing the corresponding model parameters related to examinees' knowledge states. For example, Lord et al.~\cite{lord1980applications} established a CAT system, where they proposed a maximum fisher information strategy that greedily selected the questions with minimizing the variance of examinee's parameters obtained by specific IRT model. Although these works have made great success, they are all \textbf{model-specific}, i.e., the CAT solution is only suitable for its designated CDM. In other words, we have to understand how a specific CDM (e.g., IRT or MIRT) works in detail when designing a CAT selection strategy. Therefore, existing CAT systems usually become inflexible since we must redesign the selection strategy if we replace the CDM behind. However, as we mentioned above, CAT primarily aims to select appropriate questions during testing so it gives us an intuition that we only need to care about the results of examinees' states no matter which CDM models we probe. To this end, we argue that an ideal CAT framework should be \textbf{model-agnostic}, i.e., the CAT solution can adapt to different CDMs.

To the best of our knowledge, no previous work on CAT has attempted to achieve the model-agnostic framework. Fortunately, we notice a similar idea from active learning (AL), which is a popular framework to alleviate data deficiency in many tasks including image classification, recommendation, etc.~\cite{gal2017deep,cai2019multi}. In principle, AL framework aims to design a strategy that selects valuable data step by step to experts for annotations so that machine learning models can be well trained in the supervised manner. Intuitively, AL inspires us for the CAT solution since it can also overlook which machine learning models we have to use. Therefore, in this paper, we propose a novel \textbf{M}odel-\textbf{A}gnostic \textbf{A}daptive \textbf{T}esting framework (MAAT) for CAT solution, where we take advantage of the general idea of active learning at the macro level.

In order to support MAAT procedure that evaluates the knowledge states of examinees comprehensively, we argue that there are two necessary objectives should be considered in designing the strategy for adaptively selecting questions: (1) \emph{Quality}. Primarily, selecting high-quality questions for testing can make the CAT procedure more efficiently. For example, it is inappropriate if we  always push examinees to try questions that are either too difficulty or too easy, because we cannot obtain the accurate diagnosis results about them at all. Therefore, an effective method that can evaluate the informativeness of questions is urgent but non-trivial. (2) \emph{Diversity}. In a certain domain (e.g., math), there is usually much knowledge (e.g., ``Function'') that examinees should learn. With CAT process, we are required to evaluate that how they master all the knowledge concepts. However, if we follow the traditional solutions~\cite{chang2015psychometrics,vie2017review}, the results may be suboptimal, since few of them directly consider such diversity issue, leading to very limited question selections on concepts. Therefore, it is necessary to select questions in MAAT that cover concepts as much as possible.

To address the above problem with considering both objectives above, we implement our MAAT framework with proposing three modules, i.e., Quality Module, Diversity Module and Importance Module. Specifically, at one CAT selection step, Quality Module first generates a small candidate subset of the most high-quality questions from the pool, where a novel score function is proposed to quantify the information gain of questions on knowledge mastery after the examinee has taken. Different from existing solutions, this module is flexible since it evaluates the informativeness through the Expected Model Change (EMC) without awareness of the detailed mechanism behind the change. Then, Diversity Module selects one question from candidates that maximizes the concept coverage in the whole CAT procedures. We also propose Importance Module to estimate the importance of concepts that optimizes the selection procedure. Moreover, we prove that the problem of maximizing both quality and diversity becomes NP-hard under our sophisticated coverage score function and provide efficient algorithms by exploiting the submodular property. Finally, we conduct extensive experiments on two real-world datasets. The experimental results demonstrate that our MAAT can select both high-quality and diverse questions in a model-agnostic way, which can support many CAT scenarios.

%% file: related-work.tex
\section{Related Work}
\label{section:rel}


\subsubsection{Computerized Adaptive Testing}
The development of Computerized Adaptive Testing (CAT) originates from the belief that tests can be more effective if we tailor them for examinees \cite{wainer2000computerized}.
Though there are some variants recently \cite{tu2019cognitive,lin2019item}, the primary challenge of CAT lies in designing a selection strategy which selects appropriate questions for the examinee step by step.
Since current strategies are closely bound to the underlying cognitive diagnosis models (CDM) \cite{torre2014cognitively,huang2020learning}, we review them in terms of the CDMs they base on.
Representative CDMs include traditional item response theory (IRT) family \cite{embretson2013item,ackerman2003using} and recently proposed deep learning models \cite{piech2015deep,wang2020neuralcd}.
IRT-based CAT strategies minimize the statistical estimation error of the latent parameter in IRT \cite{chang1996global,chang2015psychometrics}.
MIRT-based strategies are proposed as multivariate extensions of the IRT-based ones \cite{wang2011item,hooker2009paradoxical,rudner2002examination}.
To the best of our knowledge, little progress has been made in designing strategies for the deep learning models due to their parametric complexity.
Having made great effect though, current strategies suffer from two limitations.
First, they must understand how the underlying CDM works in detail, making CAT systems model-agnostic and inflexible.
Second, they overlook the diversity in question selection, causing potential imbalance in the diagnosis for knowledge concept mastery.
We provide novel solutions within our proposed model-agnostic framework, which will be discussed in Section~\ref{section:framework}.

\subsubsection{Active Learning}
Active learning (AL) is motivated by the belief that we can train a better model with less data if we actively select valuable data, and has been applied in many supervised learning tasks \cite{gal2017deep,cai2019multi}.
Starting with a machine learning model and a data selection strategy, at each step, AL framework selects a batch of unlabeled data to be annotated for supplementing the limited labeled data so that the model achieves better performance.
The key point is how to avoid using model details in strategy design so that it can apply to varieties of tasks with different models.
Generally, there are two solutions which utilize model outputs and data features, respectively \cite{huang2010active}.
Specifically, \textit{uncertainty} based algorithms examine the label predictions output by the model and select the ones whose predictions contain the most uncertainty \cite{gal2017deep,bachman2017learning};
\textit{representativeness} based algorithms examine the features of data samples and select the ones which represent the overall patterns of unlabeled data best \cite{ting2018optimal}.
The methodologies inspire us to propose a model-agnostic solution for CAT as well,
however with different goals dedicated to CAT (i.e., quality and diversity) and a novel strategy to optimize the goals.

\subsubsection{Coverage Measure}
The coverage measure has been extensively studied in tasks related to document summarization \cite{lin2011class,sipos2012temporal}, network analysis \cite{liu2014influence} and recommendation \cite{wu2016relevance,hammar2013using}, in which we try to find a subset that covers as much information in the document or recommendation as possible.
In many scenarios such as recommendation, it is intuitively better to consider such coverage objective \cite{huang2019exploring,wu2016relevance,wang2018united,zhang2020personalized,liu2011personalized}.
Specially, some previous works utilize \textit{submodularity} in their design of coverage score function for optimization \cite{lin2011class,liu2014influence,wu2016relevance}, which is a mathematical modeling towards the intuitive diminishing returns property.
To the best of our knowledge, our work is the first attempt to explicitly define the diversity goal of knowledge concepts in CAT with a formulated coverage measure, where we provide an efficient optimization algorithm by exploiting the inherent submodularity property.

%% file: preliminaries.tex
\section{Preliminaries}
\label{section:pre}

This section discusses the terminologies, the goals, and the reformulation of computerized adaptive testing (CAT).

\begin{figure*}[htb]
    \centering 
    \includegraphics[width=\textwidth]{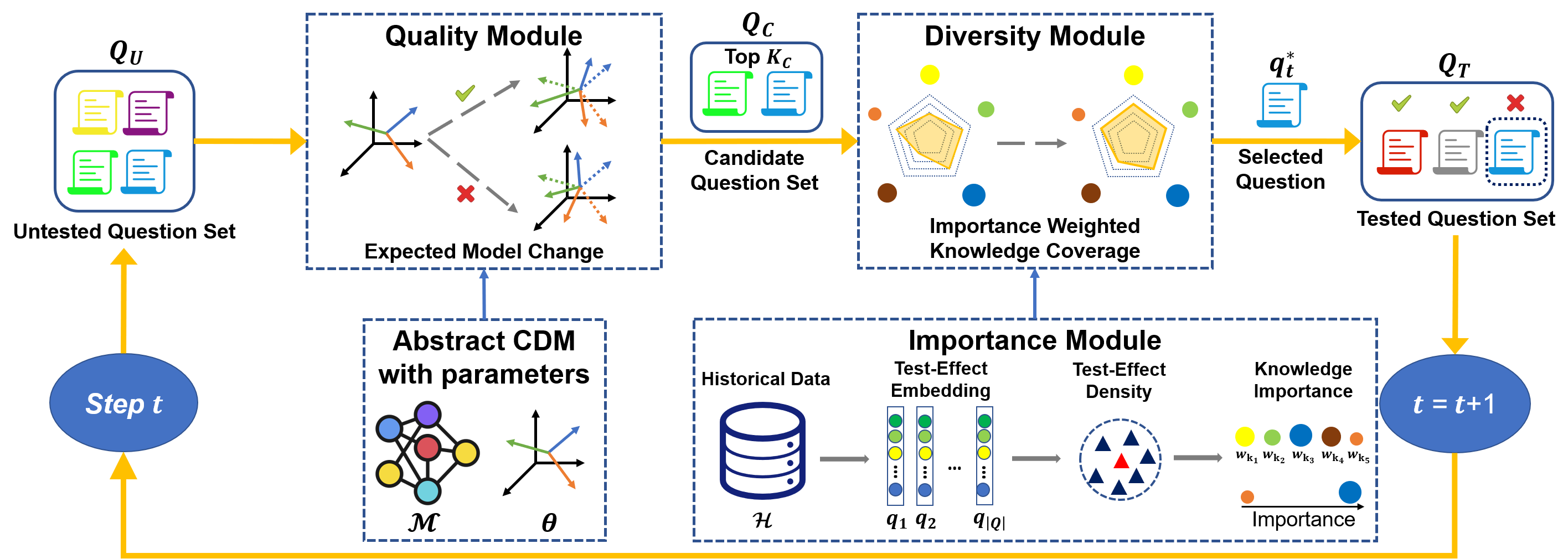}
    \caption{Overview of our MAAT framework.}
    \label{Fig.2}
\vspace{-15pt}
\end{figure*}

\subsection{Terminologies}
\subsubsection{Environment}
As a specific form of test, a CAT system works with a typical testing environment consisting of examinees and questions.
Suppose there is an examinee set $E\!=\!\left\{e_1,e_2,...,e_{|E|}\right\}$ and a question set $Q\!=\!\left\{q_1,q_2,...,q_{|Q|}\right\}$.
We denote the record of examinee $e_i$ answering question $q_j$ as a triplet $r_{ij}\! =\!<\!e_i ,q_j ,a_{ij}\!>$, where $a_{ij}$ equals 1 if $e_i$ answers $q_j$ correctly, and 0 otherwise. We denote all the records as $R$, and the records belonging to a certain examinee $e_i$ as $R_i$.
In addition, we suppose a set of knowledge concepts $K\!=\!\left\{k_1,k_2,...,k_{|K|}\right\}$ related to the questions.
We denote the association between questions and concepts as a binary relation $G \subseteq Q\times K$, where $(q_i,k_j)\in G$ if $q_i$ is related to $k_j$.

\subsubsection{Status}
Besides the static environment, a CAT system maintains some dynamic status dedicated to adaptive tests.
Specifically, within the test for a certain examinee $e_i\in E$, the question set $Q$ is divided into a tested set $Q_T$ and an untested set $Q_U$.
Initially, $Q_U\!=\!Q, Q_T\!=\!\emptyset$.
At each step, one question is selected from $Q_U$ to $Q_T$.
When the test finishes, $Q_T$ forms a tailored test sequence for $e_i$.

\subsubsection{Components}
Finally, we formulate the components of a CAT system.
Following Fig.~\ref{subfig:CATcomponents}, we formally denote a CAT system as $(\mathcal{M}, \mathcal{S})$, where $\mathcal{M}$ is a cognitive diagnosis model (CDM) and $\mathcal{S}$ is a question selection strategy. 
Different from traditional CAT systems, in our problem, $\mathcal{M}$ does not refer to any specific CDM (e.g., IRT), but an abstract model with two basic functionalities: (1) $\mathcal{M}$ captures the knowledge states of the examinees with a group of parameters $\boldsymbol{\theta}$ without any assumption about the detailed form or mechanism; (2) given an examinee $e_i\in E$ and a question $q_j\in Q$, $\mathcal{M}$ can output a performance prediction $\mathcal{M}(e_i,q_j|\boldsymbol{\theta})\in[0,1]$ which measures how likely $e_i$ can answer $q_j$ correctly.
$\mathcal{S}$ accepts $Q_U$ and $\mathcal{M}$ as input, and outputs a question $q\in Q_U$, i.e., $q=\mathcal{S}(Q_U, \mathcal{M})$. In other words, it makes the selection from the untested question set according to the current estimated knowledge states.

\subsection{Goals}

We now discuss the two goals for the selection strategy $\mathcal{S}$.

\subsubsection{Quality}
Generally, a high-quality question helps reduce the uncertainty of the examinee's knowledge states.
Therefore, we quantify the quality of a question through its \textit{informativeness}, i.e., how much information the underlying model $\mathcal{M}$ can obtain from the question to update the estimate for knowledge states.
In this way, achieving the quality goal means to select the most informative questions.
To evaluate informativeness, after the test for $e_i\!\in \!E$, we predict the her performance with $\mathcal{M}$ on the whole question pool, and measure it with some metric such as AUC.
We denote such measurement as $Inf(\mathcal{S})$, which will be discussed in detail in Section~\ref{section:exp}.

\subsubsection{Diversity}
Generally, we consider a set of questions to be diverse if it meets certain coverage requirements.
We intuitively measure diversity with knowledge coverage.
As a result, achieving the diversity goal means to select a set of questions that has the maximum knowledge concept coverage.
After the test, We can evaluate the coverage by the proportion of the knowledge concepts the tested question set $Q_T$ covers, denoted as $Cov(\mathcal{S})$. We will discuss it in detail in Section~\ref{section:exp}.

\begin{table}
  \caption{Concept correspondence between CAT and AL}
  \label{tab:catvsal}
  \begin{tabular}{c|c|c}
    \hline
    CAT Concepts & AL Concepts & Notation\\
    \hline
    Cognitive diagnosis model & Supervised learning model & $\mathcal{M}$\\
    Question selection strategy & Sample query strategy & $\mathcal{S}$ \\
    Examinees & Expert annotators & $E$ \\
    Questions & Data samples & $Q$ \\
    Tested/Untested questions & Labeled/Unlabeled samples & $Q_T, Q_U$ \\
  \hline
\end{tabular}
\vspace{-15pt}
\end{table}

\subsection{Problem Formulation}
Inspired by active learning (Table~\ref{tab:catvsal}), we reformulate our model-agnostic CAT problem, however, with a key difference that we aim to achieve the problem considering both quality and diversity goals as discussed above:

\textbf{Problem Definition.}
Given a new examinee $e_i\!\in \!E$, a question pool $Q$ with knowledge concepts $K$,
our task is to design a strategy $\mathcal{S}$ to select a $N$-size question set $Q_T\!=\!\left\{q_1^*, q_2^*, ..., q_N^*\right\}$ step by step that has the maximum quality and diversity.
Before testing, we set up an abstract CDM $\mathcal{M}$ with parameters $\boldsymbol{\theta}$ capturing knowledge states.
During testing, at step $t$ $(1\!\leq \!t\!\leq\! N)$, we select one question $q_t^*=\mathcal{S}(Q_U, \mathcal{M})$, then observe a new interaction test record $r_{it}^*=<e_i, q_t^*, a_{it}^*>$ and update the knowledge states, i.e., $\boldsymbol{\theta}$, in $\mathcal{M}$ instantly. 
After testing, we measure the effectiveness of $\mathcal{S}$ by computing $Inf(\mathcal{S})$ and $Cov(\mathcal{S})$.

%% file: MAAT.tex
\section{MAAT Framework}
\label{section:framework}


\subsection{Overview}
\label{subsection:overview}
We present the overview of our MAAT framework in Fig.~\ref{Fig.2}.
For any examinee $e_i\in E$, the proposed framework appropriately generates a question set step by step in interaction with her.
At step $t$, MAAT selects one question from the untested question set $Q_U$ to the tested quesion set $Q_T$, which optimizes both quality and diversity goals.
The overall architecture can be seen as three modules: \textit{Quality Module}, \textit{Diversity Module} and \textit{Importance Module}.
Specifically, at step $t$, in Quality Module, a candidate question set  $Q_C$ is selected from $Q_U$, which consists of the top-$K_C$ high-quality questions with maximum informativeness.
Then in Diversity Module, one final question $q_t^*$ is selected from $Q_C$ to $Q_T$, which contributes the maximum marginal gain to the knowledge coverage of $Q_T$.
Additionally, to achieve a more efficient selection procedure, we propose Importance Module to evaluate the importance of knowledge concepts, which utilizes historical data for pre-computation before testing.


\subsection{Quality Module}
\label{subsection:memc}
We first introduce Quality Module, which is the first stage of step $t$ as shown in the upper left part of Fig.~\ref{Fig.2}.
The aim of this module is to select a candidate set $Q_C$ from $Q_U$ consisting of the top-$K_C$ high-quality questions ($K_C\!=\!|Q_C|$).
To achieve that, we propose a score function, namely \textit{Expected Model Change} (EMC), which quantifies the quality of questions by measuring their informativeness, i.e., how much information they contain.
Different from traditional CAT heuristic functions \cite{chang2015psychometrics,vie2017review}, EMC is independent of the details of the CDM.

The general idea behind is as follows.
We make use of the information contained in a question by updating our estimate to the examinee's knowledge states after observing her answer.
Therefore, the informativeness can be scored by how much the diagnosis changes through a question.
In MAAT, the knowledge states are carried by the parameters of the abstract CDM, i.e., $\boldsymbol{\theta}$ in $\mathcal{M}$.
Regardless of the concrete mechanism, how much the CDM changes implies the amount of information obtained from the question.
Specifically, if $\boldsymbol{\theta}$ has a great change, the question can be considered to be informative. Otherwise, if $\boldsymbol{\theta}$ hardly changes, knowing the response to the question brings little information.

The challenge is that it is impossible to know the examinee's response before selection. Therefore, we compute the expectation of model change w.r.t. the probability that the examinee answers the question correctly, which is predicted by $\mathcal{M}$.
Formally, let $\Delta \mathcal{M}(r_{ij})=|\boldsymbol{\theta}(R_i\cup\left\{r_{ij}\right\})-\boldsymbol{\theta}(R_i)|$ be the model change that would be obtained by adding the record $r_{ij}=<e_i,q_j,a_{ij}>$, where $\boldsymbol{\theta}(R_i)$ denotes the parameters trained with $e_i$'s current records, $R_i$, and so as $\boldsymbol{\theta}(R_i\cup\left\{r_{ij}\right\})$.
For every $q_j\in Q_U$, we define its \textit{expected model change} as:
\begin{gather}
  {\rm EMC}(q_j)=\mathbb{E}_{a_{ij}\sim p}\Delta \mathcal{M}(<e_i,q_j,a_{ij}>)\label{equation:emc}, \\
  p=\mathcal{M}(e_i,q_j|\boldsymbol{\theta}(R_i)).
\end{gather}
For computational efficiency, we approximate $\Delta \mathcal{M}(r_{ij})$ with the gradient caused by $r_{ij}$ instead of retraining the model.
This approaximation is especially efficient for those CDMs using gradient-based training, such as neural models.

With the EMC score function, for each untested question $q_j\in Q_U$, we evaluate ${\rm EMC}(q_j)$ (Eq.~\eqref{equation:emc}), and then select a candidate set of the top-$K_C$ high-quality questions $Q_C$ which have the maximum informativeness.


\subsection{Diversity Module}
\label{subsection:iwkc}
Now that we have a candidate set $Q_C$ containing high-quality questions, we turn to Diversity Module, which is the second stage of step $t$ as shown in the upper right part of Fig.~\ref{Fig.2}.
In this module, we aim at selecting one question $q_t^*$ from $Q_C$ that optimizes the diversity goal.
The neccessity of this stage comes from the observation that our diagnosis will be likely one-sided if we overlook the diversity in question selection.
To achieve high diversity as well, we first propose a score function that quantifies the knowledge coverage of the tested question set $Q_T$,
and then search for an algorithm to construct $Q_T$ with maximum coverage score by adding questions step by step.
The two main challenges are how to properly construct the coverage score function and how to design the optimization algorithm.

To solve the first challenge, we begin with a Naive Knowledge Coverage (NKC) function, which simply calculates the proportion of knowledge concepts covered by the selected question set $Q_T$ to all the knowledge concepts:
\begin{gather}
    {\rm NKC}(Q_T)=\frac{\sum_{k\in K} Cov(k,Q_T)}{|K|},\label{equation:nkc} \\
    Cov(k,Q_T)=\mathbbm{1}[\exists q\in Q_T,(q,k)\in G].\label{equation:naive-cov}
\end{gather}
Although NKC is intuitive, we find it suffer from two practical drawbacks:
(1) It treats all the knowledge concepts equally and cannot distinguish their importance. For example, in a math test, if our tests focus more on ``Algebra'' than ``Geometry'', we should select more questions related to ``Algebra'' besides covering both of them.
(2) It is too strict since $Cov(k,Q_T)$ is binary depending on whether $k$ is covered by $Q_T$, which means that $Cov(k,Q_T)$ always equals to 1 as long as at least one question in $Q_T$ is related to $k$, regardless of the exact count.
This strong condition might result in the imbalance of covered knowledge. For example, choosing 9 questions related to ``Algebra'' and 1 question related to ``Geometry'' is equivalent to choosing 5 of each under this condition, however the latter is usually a better choice.

To this end, following the idea in \cite{wu2016relevance}, we add two novel features to NKC (Eq.~\eqref{equation:nkc}):
(1) to take importance of knowledge concepts into consideration, we add an importance weight $w_k$ for each knowledge concept $k\in K$;
(2) to alleviate the imbalance caused by the strict binary $Cov(k,Q_T)$ in Eq.~\eqref{equation:naive-cov}, we improve it into a soft form with incremental property, i.e., the value gradually grows from 0 to 1 as there are more and more questions related to the knowledge concept.
Formally, we proposed an advanced score function named \textit{Importance Weighted Knowledge Coverage} (IWKC):
\begin{gather}
    {\rm IWKC}(Q_T)=\frac{\sum_{k\in K}w_k\times IncCov(k,Q_T)}{\sum_{k\in K}w_k}\label{equation:iwkc}, \\
    IncCov(k,Q_T)=\frac{cnt(k,Q_T)}{cnt(k,Q_T) + 1}\label{equation:inc_cov}, \\
    cnt(k,Q_T)=\sum\nolimits_{q\in Q_T}\mathbbm{1}[(q,k)\in G]\label{equation:cnt},
\end{gather}
where $w_k$ is the added importance weight for concept $k$, which is a positive constant.
We will discuss the computation of $w_k$ in Section~\ref{subsection:imp}.
$IncCov(k,Q_T)$ is the incremental improvement to $Cov(k,Q_T)$ in Eq.~\eqref{equation:naive-cov}.
For example, when the number of questions related to $k$ is 0, 1, 2, 3, ..., $IncCov(k,Q_T)$ gradually reaches 0, 0.5, 0.67, 075, ..., respectively, while as $Cov(k,Q_T)$ discontinuously jumps from 0 to 1.

Note that though IWKC has been well constructed, the sophisticated structure still brings the second challenge, i.e., how to select a set of questions which has the maximum IWKC score in a step-by-step way.
Indeed, this optimization problem is proved to be NP-hard (Section~\ref{subsection:proof}).
Fortunately, we find a suboptimal solution with acceptable performance by exploiting the \textit{submodular} property of IWKC.
Generally, submodularity can be seen as a mathematical modelling to the narural \textit{diminishing returns} property \cite{nemhauser1978analysis}.
For a submodular set function, as the set gets larger, the marginal gain obtained by adding one more element will decrease.
Specifically, in our case, as $Q_T$ grows larger with selection steps going on, the gain in coverage (i.e., IWKC) caused by adding the same question will get slower.
For clarity of our discussion, we leave the formal proof later in Section~\ref{subsection:proof}.
The submodular property of IWKC provides us with a performance-guaranteed greedy selection algorithm \cite{nemhauser1978analysis}.
Generally, at step $t$ in the test, for each candidate question $q_j\in Q_C$, we evaluate the marginal gain of IWKC for $Q_T$ if $q_j$ were added to $Q_T$, and greedily select the one maximizing the marginal gain as the $t$-step selection $q_t^*$:
\begin{gather}
q_t^* = argmax_{q_j\in Q_C} \Delta_{q_j} {\rm IWKC}(Q_T),\label{equation:greedy_algo} \\
\Delta_{q_j} {\rm IWKC}(Q_T) = {\rm IWKC}(Q_T \cup \left\{q_j\right\}) - {\rm IWKC}(Q_T).\label{equation:iwkc_gain}
\end{gather}
With the above algorithm, the ratio of $Q_T$'s IWKC to the optimal value is guaranteed to be at least $1-\frac{1}{e}$.


\subsection{Importance Module}
\label{subsection:imp}
After demonstrating the two-stage procedure to select an optimal question at each step during testing, we turn to solve the problem of computing the importance weight $w_k$ in IWKC (Eq.~\eqref{equation:iwkc}).
As shown in the lower right part of Fig.~\ref{Fig.2}, Importance Module pre-computes $w_k$ for each knowledge concept before the test begins.
The general idea is to consider a knowledge concept to be important if its associated questions are more \textit{representative}.
Typically, a question is considered to be representative if it has similar characteristics with many other questions.
With a representative question, we can implicitly examine many questions at the same time.
To quantify the representativeness of questions, we firstly represent them with feature vectors so that each question can be seen as a point in the embedding metric space.
The closer a question is to its neighbors, the more representative the question.
Finally, we obtain the importance weight of each knowledge concept by averaging the representativenss (i.e., the density) of its related questions.
To accomplish the computation, we utilize the historical data of the CAT system. Specifically, we have historical examinees whose records are persisted and can be used for training, which we denote as $\mathcal{H}=(E^H, R^H)$.

\subsubsection{Test-Effect Embedding}
The key point of embedding questions is to define the distance metric between questions.
The general idea is that the historical examinees' performance on the question can characterize the question itself, such as difficulty and differentiation.
Thus we declare that questions on which examinees perform similarly have \textit{Test-Effect similarity}, and define the question embedding following such similarity as \textit{Test-Effect embedding}.
In order to train Test-Effect embeddings, we extend the idea from \textit{Item2Vec} \cite{barkan2016item2vec}.
Specifically, for each historical record $r_{ij}=<e_i,q_j,a_{ij}>$, we set up an input $\boldsymbol{x}_{ij}$ to represent both which question was answered and if the question was answered correctly:
\begin{equation}
\setlength{\abovedisplayskip}{3pt}
\setlength{\belowdisplayskip}{3pt}
     \boldsymbol{x}_{ij}=\left\{
    \begin{aligned}
    &\boldsymbol{1}_{|Q|}(j)\oplus \boldsymbol{1}_{|Q|}(j), & \text{if $a_{ij}=1$} \\
    &\boldsymbol{1}_{|Q|}(j)\oplus \boldsymbol{0}_{|Q|}, & \text{if $a_{ij}=0$}
    \end{aligned}
     \right.,
\end{equation}
where $\boldsymbol{1}_{|Q|}(j)$ denotes a $|Q|$-length one-hot vector with only the $j$th position equal to 1, $\boldsymbol{0}_{|Q|}$ denotes a $|Q|$-length zero vector, and $\oplus$ denotes vector concatenation. Then we train a \textit{Skip-Gram Negative Sampling} (SGNS) model \cite{mikolov2013distributed}. Specifically, given a historical examinee $e_k\in E^H$, the optimization objective is formulated as:
\begin{small}
\begin{gather}
    \max \frac{1}{|R^H_k|}\sum_{r_{ki}\in R^H_k}\sum_{r_{kj}\in R^H_k, j\neq i} \log p(r_{kj}|r_{ki}),\label{equation:embedding1} \\
    p(r_{kj}|r_{ki})=\sigma((\boldsymbol{W}\boldsymbol{x}_{ki})^T\boldsymbol{v}_j)\prod_{l=1}^{N_{neg}}\sigma(-(\boldsymbol{W}\boldsymbol{x}_{kn_l})^T\boldsymbol{v}_{n_l}),\label{equation:embedding2}
\end{gather}
\end{small}
where $N_{neg}$ is the negative sampling size, $\sigma(x)=\frac{1}{1+e^{-x}}$ is the Sigmoid function, $\boldsymbol{W}$ is a $d\times 2|Q|$ parameter matrix,
and $\boldsymbol{v}_j$ is our Test-Effect embedding of $q_j$ with dimension $d$, which we denote as $\boldsymbol{q}_j^{TE}$.

\subsubsection{Test-Effect Density}
Since the questions have been represented in the Test-Effect embedding space, we quantify the representativeness of questions.
First, We compute Test-Effect similarity between two questions $q_i$ and $q_j$ as
\begin{equation}
\setlength{\abovedisplayskip}{3pt}
\setlength{\belowdisplayskip}{3pt}
    Sim_{TE}(q_i,q_j)=e^{-\gamma|\boldsymbol{q}_i^{TE}-\boldsymbol{q}_j^{TE}|}, \label{equation:te-sim}
\end{equation}
where $|\cdot|$ is Euclidean norm and $\gamma$ is a positive smoothing parameter.
Next, we define the \textit{Test-Effect density} of each question $q_j$ as the average similarity to its neighbors:
\begin{equation}
\setlength{\abovedisplayskip}{3pt}
\setlength{\belowdisplayskip}{3pt}
    Den_{TE}(q_j)=\frac{1}{K_N}\sum_{q_i\in \mathcal{N}(q_j)}Sim_{TE}(q_i,q_j), \label{equation:te-dens}
\end{equation}
where $\mathcal{N}(q_j)$ contains the $K_N$-nearest-neighbors of $q_j$ in the Test-Effect embedding space. The larger Test-Effect density a question has, the more representative it is.

\subsubsection{Knowledge Importance}
At last, we define the knowledge importance of each knowledge concept $k$, i.e., $w_k$, by averaging the Test-Effect density of the questions related to it:
\begin{equation}
\setlength{\abovedisplayskip}{3pt}
\setlength{\belowdisplayskip}{3pt}
    w_k=\frac{1}{\sum_{(q,k)\in G}{1}}\sum_{(q,k)\in G}Den_{TE}(q). \label{equation:importance}
\end{equation}
Substituting Eq.~\eqref{equation:importance} into Eq.~\eqref{equation:iwkc}, we get the complete IWKC.


\subsection{Theoretical Analysis}
\label{subsection:proof}
To be rigorous, we supplement the theoretical proofs about the optimization with IWKC score function (Section~\ref{subsection:iwkc}).
Formally, we define the \textit{IWKC Maximization Problem} as follows:
given a question set $Q$ with associated knowledge concepts $K$, the target is to identify a $N$-size subset $Q_T$ with the maximum IWKC:
\begin{equation}
\setlength{\abovedisplayskip}{3pt}
\setlength{\belowdisplayskip}{3pt}
    \max_{Q_T\subset Q, |Q_T|=N}{\rm IWKC}(Q_T).
\end{equation}

First, we demonstrate the complexity of this problem:

\begin{theorem}
The IWKC maximization problem is NP-hard.
\end{theorem}

\begin{IEEEproof}
First, we introduce a classic NP-hard question, namely \textit{weighted maximum coverage problem}:
given a number $k$, a collection of sets $S\!=\!\left\{S_1,S_2,...,S_m\right\}$, a domain of elements $E\!=\!\left\{e_1,e_2,...,e_n\right\}$ each of which has a weight $w_i,i\!=\!1,...,n$, the objective is to find a subset $S_{opt}\subset S$ such that $|S_{opt}|\le k$ and the total weights of the elements covered by $\bigcup_{S_i\in S_{opt}}S_i$ is maximized.
Next, we reduce the IWKC maximization problem to the weighted maximum coverage problem.
We consider the knowledge concepts as the corresponding elements, and the question set $Q\!=\!\left\{q_1,q_2,...,q_m\right\}$ covering concepts as the corresponding collection of sets.
The weight of each concept $k$ corresponds to $w_k\times IncCov(k,Q)$ as defined in Eq.~\eqref{equation:iwkc}.
Under this situation, the IWKC maximization problem is equivalent to the weighted maximum coverage problem, and therefore is NP-hard.
\end{IEEEproof}

Then we verify the submodular property of IWKC:

\begin{theorem}
IWKC is a nonnegative monotone submodular function.
\end{theorem}
    
\begin{IEEEproof}
The nonnegativity and monotonicity of IWKC are obvious since it grows from 0 to 1, so we focus on proving the submodularity.
First, we claim that $IncCov(k,Q_T)$ (Eq.~\eqref{equation:inc_cov}) is submodular.
Let $\Delta_{q_j} IncCov(k,Q_T)\!=\!IncCov(k,Q_T\cup\left\{q_j\right\}) - IncCov(k,Q_T)$ be the marginal gain of $IncCov(k,Q_T)$ when $q_j$ is added, which is calculated as $\Delta_{q_j} IncCov(k,Q_T)\!=\!\frac{cnt(k,Q_T\cup\left\{q_j\right\})-cnt(k,Q_T)}
{(cnt(k,Q_T\cup\left\{q_j\right\}) + 1)(cnt(k,Q_T) + 1)}$.
Consider the tested question sets at two consecutive steps, $Q_T'\subset Q_T''\subset Q$.
As $Q_T'\subset Q_T''$, we have
$cnt(k,Q_T'\cup\left\{q_j\right\})-cnt(k,Q_T') \ge
cnt(k,Q_T''\cup\left\{q_j\right\})-cnt(k,Q_T'')
$ and
$(cnt(k,Q_T'\cup\left\{q_j\right\}) + 1)(cnt(k,Q_T') + 1) \le
(cnt(k,Q_T''\cup\left\{q_j\right\}) + 1)(cnt(k,Q_T'') + 1)$.
Thus $\Delta_{q_j} IncCov(k,Q_T') \ge \Delta_{q_j} IncCov(k,Q_T'')$, which is the definition of submodularity.
Since IWKC is nonnegative linear combinations of $IncCov$, the submodularity of IWKC can be easily derived from the submodularity of $IncCov$.
\end{IEEEproof}

Finally, we review the performance guarantee of the marginal gain based greedy algorithm \cite{nemhauser1978analysis}:

\begin{theorem}
For any nonnegative monotone submodular function $F$, let $S^*$ be the N element set with the best performance and $S$ the same size set obtained by greedy algorithm, which selects an element with maximum marginal gain each time, and then $F(S)\geq(1-\frac{1}{e})F(S^*)$.
\end{theorem}
In our case, IWKC corresponds to the $F$ above, and the selected question set $Q_T$ corresponds to the $S$.


\begin{algorithm}
\caption{MAAT Flowchart}
\label{algorithm:memc}
\begin{algorithmic}
\STATE{\textbf{Data}: $Q, K, G, \mathcal{H}$}
\STATE{\textbf{Input}: $\mathcal{M}$; $e_i\in E$}
\STATE{\textbf{Output}: A $N$-length test sequence $Q_T=\left\{q_1^*,q_2^*,...,q_N^*\right\}$}
\STATE{\textbf{Initialization}: \\
    \quad create $\mathcal{M}$ with $\mathcal{H}$ and randomly initialize $\boldsymbol{\theta}$ \\
    \quad train Test-Effect embedding for $q\in Q$ (Eq.~\eqref{equation:embedding1}~\eqref{equation:embedding2})\\
    \quad compute importance weight for $k\in K$ (Eq.~\eqref{equation:importance}) \\
    \quad $Q_U=Q, Q_T=\emptyset$\;}
\FOR{$t=1$ \TO $N$} 
    \STATE{$\forall q_j\in Q_U$, calculate ${\rm EMC}(q_j)$} (Eq.~\eqref{equation:emc})
    \STATE{$Q_C=\left\{\text{top $K_C$ questions with maximum EMC}\right\}$}
    \STATE{$q_t^*=argmax_{q_j\in Q_C} \Delta_{q_j} {\rm IWKC}(Q_T)$ (Eq.~\eqref{equation:iwkc_gain}~\eqref{equation:iwkc})}
    \STATE{$Q_U=Q_U-\left\{q_t^*\right\}$}
    \STATE{$Q_T=Q_T\cup\left\{q_t^*\right\}$} 
    \STATE{observe $r_{it}^*=<e_i,q_t^*,a_{it}^*>$}
    \STATE{update $\mathcal{M}(\boldsymbol{\theta})$}
\ENDFOR
\end{algorithmic}
\end{algorithm}

\subsection{Summary}
\label{subsection:opt}
In summary, the flowchart of MAAT framework is as Algorithm~\ref{algorithm:memc},
which adpots a two-stage solution for the multi-objective optimization \cite{zhang2020personalized,wang2018united}.
MAAT has the following advantages:
(1) it is model-agnostic and suitable for a wide range of CDMs, including those which traditional CAT methods cannot fit in;
(2) it optimizes both quality and diversity with novel score functions and has an efficient performance-guaranteed optimization algorithm.
It is worth noting that MAAT keeps a balance of the two goals with the hyperparameter $K_C$, the size of the candidate set connecting Quality Module and Diversity Module, which will be explored further in Section~\ref{section:exp}.

%% file: experiment.tex
\section{Experiment}
\label{section:exp}

In this section, we evaluate our MAAT framework on two real-world datasets.
In addition, we conduct an ablation study on how MAAT keeps a balance of quality and diversity.
Our codes are available in https://github.com/bigdata-ustc/MAAT.


\subsection{Dataset Description}
We use two real-world educational datasets, namely EXAM and ASSIST.
The EXAM dataset was supplied by iFLYTEK Co., Ltd., collected from an online educational system where students took exams for testing.
On this system, we collected the data records of junior high school students on math tests as well as the associated knowledge concepts of those questions, such as ``Algebra''.
ASSIST is an open dataset, namely \textit{Assistments 2009-2010 skill builder}\footnote{https://sites.google.com/site/assistmentsdata/}, that recorded students' practice on math.
Each question contained in ASSIST is associated with one or more knowledge concepts such as ``Absolute Value''.


\subsection{Experimental Setup}
\subsubsection{Data Preprocessing and Partition}
For the sake of the reliability the experimental results, we apply the following data preprocessing.
First, in both EXAM and ASSIST, we filter the knowledge concepts that have less than 10 related questions;
Second, in ASSIST, we filter the questions that are answered by less than 50 students and the students that answer less than 10 questions. 
Detailed statistics are presented in Table \ref{tab:data}.

In our experiment, we partition the data into \textit{historical data} and \textit{testing data} for different purposes.
The historical data, denoted as $\mathcal{H}=(E^H,R^H)$ before, are assumed known before the tests begin.
$\mathcal{H}$ is used for the CDMs to initially learn some parameters fixed during testing, such as the difficulty of the questions.
In addition, MAAT utilizes $\mathcal{H}$ as input for question embedding to compute the knowledge importance weights (Section~\ref{subsection:imp}).
The testing data are used to simulate an adaptive testing environment.
The students in testing data are treated as examinees that are new to the CAT system,
and their records are assumed unknown until we select the questions for them during testing.
On the other hand, for evaluation, we limit our selection to those questions whose response has been recorded in the testing data during our experiment.
Therefore, to ensure that the candidate question set is large enough, we partition those students with more records into the testing data.
Specifically, for EXAM, we divide the students who answered at least 100 questions into the testing data;
for ASSIST, we divide the students who answered at least 150 questions into the testing data.
The remaining parts are left as historical data.

\subsubsection{Parameter Setting}
We set the test length $N\!=\!50$, which is quite enough for typical tests in practical.
In Quality Module, we set the size of the output candidate set, $K_C\!=\!10$.
In Importance Module, we set $N_{neg}\!=\!10$ (Eq.~\eqref{equation:embedding2}), $\gamma\!=\!0.1$ (Eq.~\eqref{equation:te-sim}), $K_N\!=\!10$ (Eq.~\eqref{equation:te-dens}).

\begin{table}
  \caption{Statistics of the datasets}
  \label{tab:data}
  \centering
  \begin{tabular}{l|cc}
    \hline
    Dataset & EXAM & ASSIST \\
    \hline
    Num. students & 4,307 & 1,505 \\
    Num. questions & 527 & 932 \\
    Num. concepts & 31 & 22 \\
    Num. records & 105,586 & 59,500 \\
    Avg. records per student & 24.5 & 39.5 \\
    Avg. records per question & 200.4 & 63.8 \\
    Avg. questions per concept & 17.0 & 44.38 \\
  \hline
\end{tabular}
\vspace{-10pt}
\end{table}

\begin{table*}
\caption{Quality Comparison with informativeness metric}
\label{tab:inf_results}
\begin{minipage}[b]{0.5\textwidth}
\centering
\textbf{(a) EXAM}
\begin{tabular}{c|c|c|c|c|c|c}
\hline
    \multirow{2}{*}{Methods} & \multicolumn{2}{c|}{IRT} & \multicolumn{2}{c|}{MIRT} & \multicolumn{2}{c}{NCDM}  \\
    \cline{2-7}
    ~ & @25 & @50 & @25 & @50 &@25 & @50 \\
\hline
    RAND & 0.6435 & 0.7076 & 0.7426 & 0.7767 & 0.7081 & 0.7566 \\
\hline
    MFI & 0.7092 & 0.7207 & - & - & - & - \\
\hline
    KLI & 0.7081 & 0.7257 & - & - & - & - \\
\hline
    D-Opt & - & - & 0.7515 & 0.7710 & - & - \\
\hline
    MKLI & - & - & 0.7502 & 0.7747 & - & - \\
\hline
    \textbf{MAAT} & \textbf{0.7192} & \textbf{0.7319} & \textbf{0.7600} & \textbf{0.7861} & \textbf{0.7614} & \textbf{0.7868} \\
\hline
\end{tabular}
\end{minipage}%
\begin{minipage}[b]{0.5\textwidth}
\centering
\textbf{(b) ASSIST}
\begin{tabular}{c|c|c|c|c|c|c}
\hline
    \multirow{2}{*}{Methods} & \multicolumn{2}{c|}{IRT} & \multicolumn{2}{c|}{MIRT} & \multicolumn{2}{c}{NCDM}  \\
    \cline{2-7}
    ~ & @25 & @50 & @25 & @50 &@25 & @50 \\
\hline
    RAND & 0.6619 & 0.6664 & 0.6734 & 0.6902 & 0.6832 & 0.7217 \\
\hline
    MFI & 0.6659 & 0.6691 & - & - & - & - \\
\hline
    KLI & 0.6658 & 0.6692 & - & - & - & - \\
\hline
    D-Opt & - & - & 0.6832 & 0.7004 & - & - \\
\hline
    MKLI & - & - & 0.6781 & 0.6877 & - & - \\
\hline
    \textbf{MAAT} & \textbf{0.6674} & \textbf{0.6703} & \textbf{0.6903} & \textbf{0.7063} & \textbf{0.7084} & \textbf{0.7334} \\
\hline
\end{tabular}
\end{minipage}
\vspace{-15pt}
\end{table*}


\subsection{Baseline Approaches}
To evaluate our model-agnostic framework, we compare it with classic model-specific methods designated to two different CDMs. 
Additionally, we conduct experiments with a deep learning CDM that classic approaches do not fit in.

First, the random selection strategy, \textit{RAND}, is a benchmark to quantify the improvement of other methods.

\textbf{IRT} \cite{embretson2013item} is the most popular CDM in CAT, and the corresponding CAT baselines are:
\begin{itemize}
\item
\textit{MFI}: \textit{Maximum Fisher Information} \cite{lord1980applications},\cite{chang2015psychometrics} is the most popular selection strategy which measures the information of questions with the Fisher information function.
\item
\textit{KLI}: \textit{Kullback-Leibler Information} \cite{chang1996global} is a global information heuristic that measures the informativeness with Kullback-Leibler divergence.
\end{itemize}

\textbf{MIRT} \cite{ackerman2003using}, as a multidimensional extension of IRT, shows its potential in multitrait ability estimation. In order to adapt to MIRT, the IRT-based methods were also extended. So we compare MAAT to the following baselines on MIRT:
\begin{itemize}
\item
\textit{D-Opt}: \textit{D-Optimality} \cite{hooker2009paradoxical}, termed in the optimization termilogy, is a multivariate extension of MFI.
\item
\textit{MKLI}: \textit{Multivariate Kullback-Leibler Information} \cite{rudner2002examination} is a direct generlization of its unidimensional version, KLI.
\end{itemize}

\textbf{NCDM} (\textit{Neural Cognitive Diagnosis Model}) \cite{wang2020neuralcd} is one of the most recent deep learning CDMs.
Though NCDM has shown great power, to the best of our knowledge, there is no existing methods able to work with it because of its extremely complex mechanism.
So we compare MAAT with only RAND to show our improvement.
The point is that with a comparison between the results of MAAT on different CDMs, we can validate the advantage of being model-agnostic.


\subsection{Evaluation Metrics}
We measure quality and diversity with the informativeness and coverage of the strategy (i.e., $Inf(\mathcal{S})$ and $Cov(\mathcal{S})$) respectively (Section~\ref{section:pre}).
In addition, we introduce a metric that has been commonly used in traditional CAT studies.

\subsubsection{Informativeness Metric}
Following the discussion in Section~\ref{section:pre}, we measure the quality through the informativeness of the selection strategy.
Specifically, for each examinee $e_i$ in the testing data, we predict her performance on every question $q_j$ whose ground truth has been recorded.
Then we adopt the common AUC (Area Under ROC) metric:
\begin{equation}
\setlength{\abovedisplayskip}{3pt}
\setlength{\belowdisplayskip}{3pt}
\label{equation:inf_metric_1}
    Inf(\mathcal{S})=AUC(\left\{M(e_i,q_j|\boldsymbol{\theta})|e_i\in E, q_j\in Q\right\}).
\end{equation}

\subsubsection{Coverage Metric}
We measure the diversity through the coverage of the selection strategy.
Since there is no universal standard, we adopt a simple form for $Cov(\mathcal{S})$, i.e., the proportion of knowledge concepts covered in the questions selected by the strategy:
\begin{equation}
\label{equation:cov_metric}
\setlength{\abovedisplayskip}{3pt}
    Cov(\mathcal{S})=\frac{1}{|K|}\sum\nolimits_{k\in K}\mathbbm{1}[k\in Q_T].
\end{equation}

\subsubsection{Simulated Estimate Error Metric}
Traditional CAT studies have a different evaluation process called \textit{simulation study}, which generates imaginary data instead of using real-world data.
For example, with IRT model, a group of \textit{simulated parameters} is generated representing the ability of examinees, the difficulty of questions, etc.
These parameters are used to generate imaginary records with Item Response Theory \cite{embretson2013item}.
Then experimentally estimate the simulated parameters in turn with the records.
To evaluate the effectiveness of the strategy, at each step in the test, we calculate the mean squared error between the estimated parameters and the simulated parameters, namely \textit{Simulated Estimate Error} (SEE):
\begin{equation}
\label{equation:see_metric}
\setlength{\abovedisplayskip}{3pt}
\setlength{\belowdisplayskip}{3pt}
    SEE(\mathcal{S})= \frac{1}{|E|}\sum\nolimits_{e_i\!\in\!E}(\boldsymbol{\theta}_i - \boldsymbol{\theta}_i^*)^2,
\end{equation}
where $\boldsymbol{\theta}_i$ and $\boldsymbol{\theta}_i^*$ are the estimated and simulated parameters related to the examinee $e_i\in E$, respectively.
In traditional case studies, $\boldsymbol{\theta}_i^*$ is randomly generated as ground truth.
To show that our measurement is consistent with the traditional one, we also evaluate the proposed framework with SEE metric.
Because we conduct experiment on real-world data, we use the estimated parameters trained on the whole data records as $\boldsymbol{\theta}_i^*$, instead of generating them.


\subsection{Experimental Results}

\subsubsection{Quality Comparison}
Table \uppercase\expandafter{\romannumeral3} reports the comparison of quality with the informativeness metric, i.e., AUC@t (Eq.~\eqref{equation:inf_metric_1}).
We show the results in the middle (step $t\!=\!25$) and the end (step $t\!=\!50$) of tests.
First, MAAT is proven to be model-agnostic since it can adapt to all the CDMs.
Second, we can easily see that, on both datasets, MAAT shows outperforming results with all CDMs during the test, which indicates that MAAT is fairly effective in achieving the quality goal.
It is worth noting that MAAT makes no use of the details of the CDMs compared with the baseline approaches.
Third, the overall results become better as the CDM becomes more complex, which is reasonable because complex CDMs do better in capturing the ability of the examinees.
This observation confirms the advantage of being model-agnostic: MAAT can improve the CAT system by making it flexible to replace the CDM without redesigning the strategy.

\subsubsection{Diversity Comparison}
\begin{figure}
\centering
\subfigbottomskip=1pt
\subfigcapskip=-3pt
\subfigure[IRT on EXAM]{    
    \includegraphics[width=0.45\linewidth]{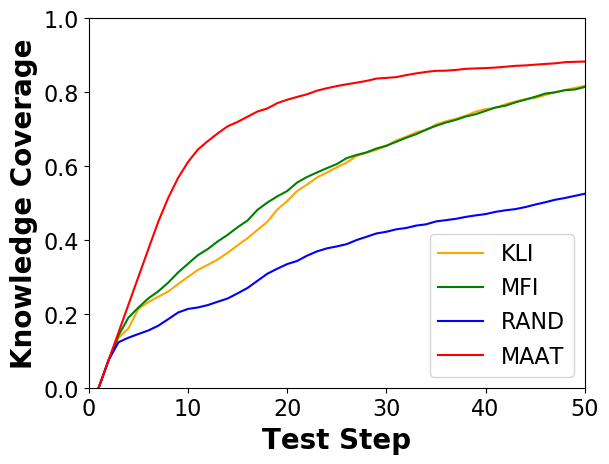}
}
\subfigure[IRT on ASSIST]{    
    \includegraphics[width=0.45\linewidth]{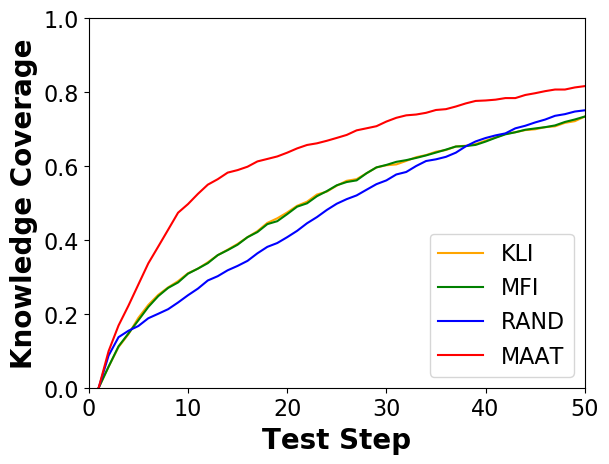}
}
\subfigure[MIRT on EXAM]{    
    \includegraphics[width=0.45\linewidth]{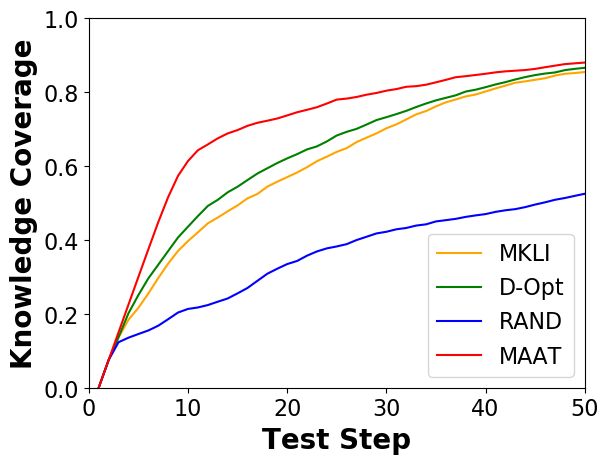}
}
\subfigure[MIRT on ASSIST]{    
    \includegraphics[width=0.45\linewidth]{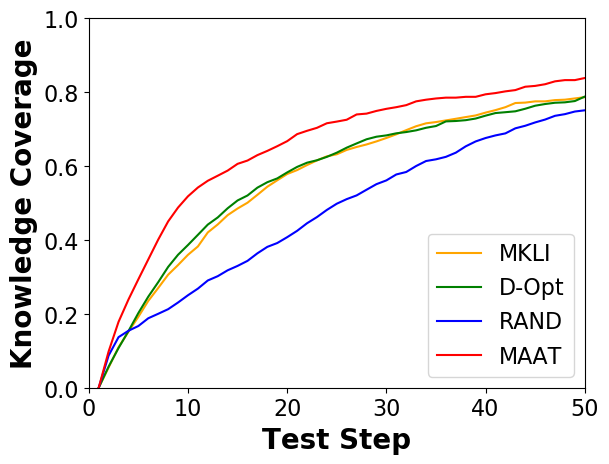}
}
\subfigure[NCDM on EXAM]{    
    \includegraphics[width=0.45\linewidth]{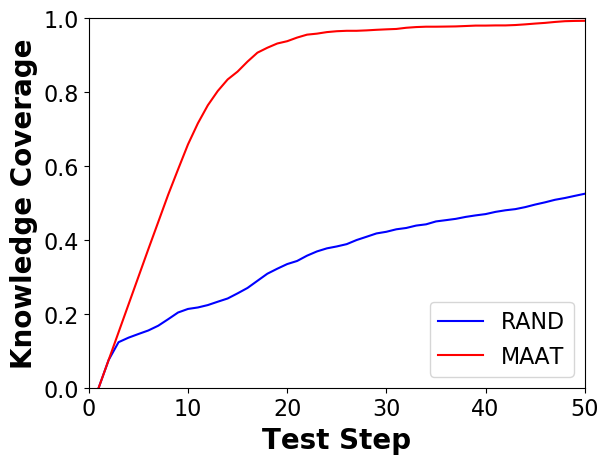}
}
\subfigure[NCDM on ASSIST]{    
    \includegraphics[width=0.45\linewidth]{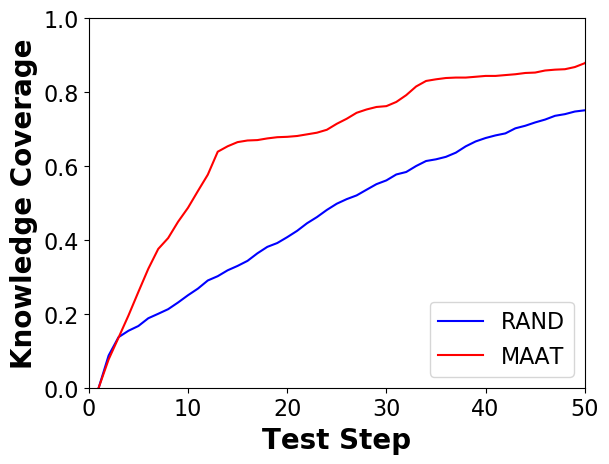}
}
\caption{Diversity Comparison with Coverage Metric}
\label{fig:cov_results}
\vspace{-10pt}
\end{figure}

Fig.~\ref{fig:cov_results} illustrates the comparison of diversity with coverage metric (Eq.~\eqref{equation:cov_metric}).
MAAT framework outperforms much on both datasets with all CDMs, because it has an intrinsic knowledge-level coverage goal and a performance-guaranteed optimization algorithm while other methods do not.
As shown in the curve charts, the coverage of MAAT grows fairly rapidly in the early steps of tests and quickly approaches the limit of 1. This feature is very important for CAT because adaptive tests are typically short.
In addition, we observe that traditional selection strategies, such as MFI and KLI, also help with the coverage goal, though they only intrinsically aim at informativeness.
This observation reveals that quality and diversity are correlated instead of contradictory.
Both of them can benefit the target of offering better diagnosis results for examinees.

\subsubsection{Consistency Validation}
\begin{figure}
\subfigbottomskip=1pt
\subfigcapskip=-3pt
\centering
\subfigure[EXAM]{    
    \includegraphics[width=0.45\linewidth]{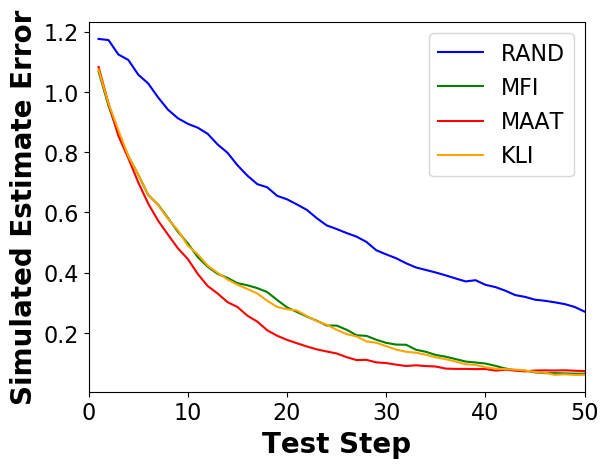}
}
\subfigure[ASSIST]{    
    \includegraphics[width=0.45\linewidth]{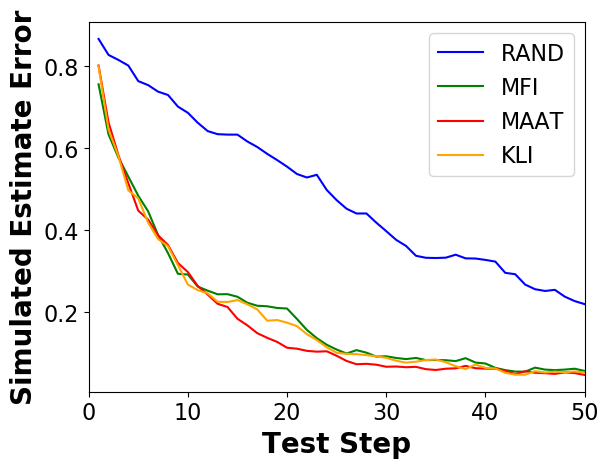}
}
\caption{Simulated Estimate Error comparison}
\label{fig:see_result}
\end{figure}

Though we have evaluated the quality and diversity of MAAT, it remains important to validate the consistency of our evaluation measurement.
Therefore, we conduct experiment with the Simulated Squared Error metric (Eq.~\eqref{equation:see_metric}) additionally.
The results are reported in Fig.~\ref{fig:see_result}.
Since simulation study is only suitable for those CDMs with extremely simple and  explainable parameters, we only conduct the simulation with IRT.
We can see that MAAT also performs well in SEE metric.

\subsubsection{Ablation Study}
\begin{figure}
\subfigbottomskip=1pt
\subfigcapskip=-3pt
\centering
\subfigure[Quality]{    
    \includegraphics[width=0.45\linewidth]{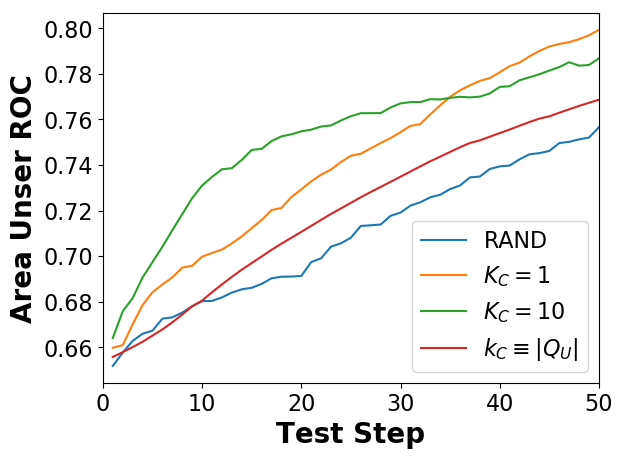}
}
\subfigure[Diversity]{
    \includegraphics[width=0.45\linewidth]{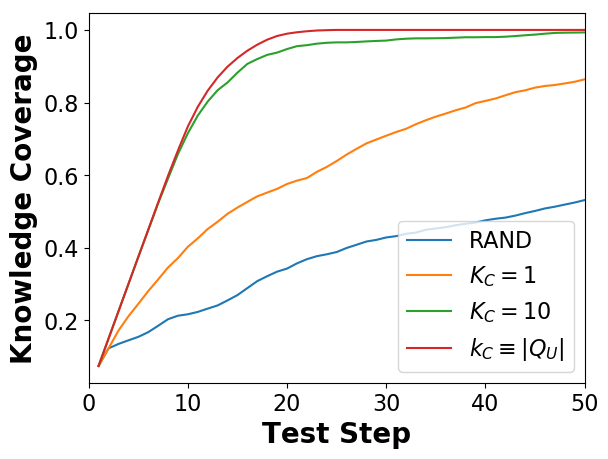}
}
\caption{Change in quality and diversity with different $K_C$ for ablation study}
\label{fig:ablation_result}
\vspace{-10pt}
\end{figure}

We conduct an ablation study on how the MAAT framework keeps a balance when quality meets diversity.
The key point is the size of the candidate set, $K_C$, which connects Quality Module and Diversity Module.
Specifically, when $K_C\!=\!1$, only Quality Module works; when $K_C\!\equiv\!|Q_U|$, only Diversity Module works.
Therefore, we observe how quality and diversity change with $K_C$ in these boundary conditions.
Due to limited pages, we show the results with NCDM on EXAM  as an example in Fig.~\ref{fig:ablation_result}.
The change in diversity is straightforward: the larger $K_C$, the faster the coverage metric increases.
Specially, when only Diversity Module works ($K_C\!\equiv\!|Q_U|$), we achieve the performance guarantee discussed in Section~\ref{subsection:iwkc}.
Moreover, we observe that the diversity quite approaches the theoretical limit when $K_C$ is a small value (i.e., $K_C\!=\!10$).
Note again that when only Quality Module works (i.e., $K_C\!=\!1$), there is still improvement on diversity compared with RAND benchmark, because the two goals are correlated.
The change in quality is slightly more interesting.
The case $K_C\!=\!1$ does not always perform the best because Diversity Module and Importance Module can also help with quality by taking coverage and importance into consideration.
To sum up, a relatively small $K_C$ keeps the best balance of quality and diversity.

\begin{table}[t]
    \caption{Results on a typical examinee for case study}
    \label{tab:case_study}
    \centering
    \begin{tabular}{c|c|c|c|c|c}
    \hline
        \multicolumn{2}{c|}{MAAT} & \multicolumn{2}{c|}{D-Opt} & \multicolumn{2}{c}{MKLI} \\
    \hline
        Concept & Inf & Concept & Inf & Concept & Inf \\
    \hline
        Function & 0.6666 & Function & 0.6652 & Triangle & 0.6645 \\
    \hline
        Set & 0.6710 & Equation & 0.6686 & Algebra & 0.6689 \\
    \hline
        Equation & 0.6763 & Equation & 0.6717 & Equation & 0.6732 \\
    \hline
        Triangle & 0.6841 & Triangle & 0.6756 & Function & 0.6774 \\
    \hline
        Algebra & 0.6905 & Geometry & 0.6801 & Algebra & 0.6810 \\
    \hline
        Triangle & 0.6961 & Function & 0.6857 & Function & 0.6843 \\
    \hline
        Coordinates & 0.7022 & Geometry & 0.6914 & Function & 0.6887 \\
    \hline
        Geometry & 0.7087 & Triangle & 0.6956 & Triangle & 0.6929 \\
    \hline
        Real Number & 0.7136 & Algebra & 0.6963 & Inequality & 0.7001 \\
    \hline
        Equation & 0.7188 & Function & 0.6998 & Geometry & 0.7057 \\
    \hline
    \end{tabular}
\vspace{-10pt}
\end{table}

\subsubsection{Case Study}
We present the first 10 steps of a typical examinee in EXAM for case study (Table~\ref{tab:case_study}).
For better illustration, we only compare MAAT with the best baselines in the previous experiment, i.e., D-Opt and MKLI on MIRT.
For each methods, we show the associated knowledge concepts in the first column and the informativeness metric, i.e. AUC@t (Eq.~\eqref{equation:inf_metric_1}), at the corresponding step in the second column.
We abbreviate the knowledge concepts, such as ``Linear Equation in One Variable'' to ``Equation''.
We can clearly see that, our MAAT framework can select diverse questions while keeping high quality as well.
Specifically, with a slightly outperforming AUC, MAAT covered 9 knowledge concepts with the first 10 selected questions, while D-Opt and MKLI covered only 5 and 6 respectively. Moreover, the two baselines tended to select more questions about ``Function'', while MAAT did not.
Therefore, our MAAT framework makes better selections with a good balance of quality and diversity.

%% file: conclusion.tex
\section{Conclusions}

In this paper, we studied a novel model-agnostic CAT problem.
We proposed a novel \textit{M}odel-\textit{A}gnostic \textit{A}daptive \textit{T}esting framework (MAAT) for the solution with addressing the problem of selecting both high-quality and diverse questions in the testing procedure.
In MAAT, we designed three sophisticated modules that worked cooperatively and iteratively.
At each selection step, Quality Module firstly selected a small candidate set of the most informative questions with EMC score function.
Diversity Module then selected one question from the candidates maximizing the knowledge coverage via IWKC, where the importance weights were pre-computed in Importance Module.
Moreover, we proved that our problem was NP-hard, and provided an efficient and effective solution by the submodular property.
Extensive experiments demonstrated that MAAT was flexible for any CDMs and could generate both high-quality and diverse questions in CAT.
We hope this work can lead to more studies in the future.

%% file: acknowledgement.tex
\section{Acknowledgement}

This research was partially supported by grants from the National Key Research and Development Program of China (No. 2016YFB1000904), the National Natural Science Foundation of China (No.s 61922073 and 61727809), and the Iflytek joint research
program.
Haiping Ma gratefully acknowledges the support of the CCF-Tencent Open Research Fund.